# Expectation Propagation for Continuous Time Bayesian Networks

**Uri Nodelman**
Stanford University
nodelman@cs.stanford.edu

**Daphne Koller**
Stanford University
koller@cs.stanford.edu

**Christian R. Shelton**
UC Riverside
cshelton@cs.ucr.edu

## Abstract

Continuous time Bayesian networks (CTBNs) describe structured stochastic processes with finitely many states that evolve over continuous time. A CTBN is a directed (possibly cyclic) dependency graph over a set of variables, each of which represents a finite state continuous time Markov process whose transition model is a function of its parents. As shown previously, exact inference in CTBNs is intractable. We address the problem of approximate inference, allowing for general queries conditioned on evidence over continuous time intervals and at discrete time points. We show how CTBNs can be parameterized within the exponential family, and use that insight to develop a message passing scheme in cluster graphs and allows us to apply expectation propagation to CTBNs. The clusters in our cluster graph do not contain distributions over the cluster variables at individual time points, but distributions over trajectories of the variables throughout a duration. Thus, unlike discrete time temporal models such as dynamic Bayesian networks, we can adapt the time granularity at which we reason for different variables and in different conditions.

## 1 Introduction

Many applications involve reasoning about a complex system that evolves over time. A standard approach is to discretize time at fixed intervals, known as time slices, and then model the system as evolving discretely from one time slice to the next. Observations are only incorporated as evidence at these fixed time points, and queries can only be asked about the system state at these times.

A time slice model is very appropriate in many applications, e.g., those where evidence is obtained at regular intervals from some sensor. However, other settings are better modeled using a less rigid notion of temporal evolution. In many systems, there is no natural time granularity: Some variables evolve quickly, whereas others change more slowly; even the same variable can change quickly in some conditions and slowly in others. Second, our ability to observe the system can vary significantly over time. We might have stretches where a variable is not observed at all, and intervals where we observe its entire trajectory; in other settings, we might only obtain evidence about certain state transitions (e.g., a marriage, a birth, a graduation). Attempts to model such systems as evolving over uniform discrete time intervals leads to very coarse approximations, or requires the entire trajectory be modeled at a very fine granularity, at high computational cost.

An alternative approach is to model such systems as evolving over continuous time, e.g., as a Markov process (Duffie et al., 1996; Lando, 1998). Nodelman et al. (2002) (NSK from now on) extend Markov processes for factored domains, defining *continuous time Bayesian networks (CTBNs)* — a structured representation for complex systems evolving over continuous time. A CTBN encodes a homogeneous continuous-time Markov process over an exponentially large state space, consisting of the assignments to a set of variables.

Exact inference in CTBNs involves generating a single matrix representing the transition model over the entire system state. As the number of states is exponential in the number of variables, this approach is generally intractable. NSK describe an approximate inference algorithm based on ideas from clique tree inference, but provide no formal justification for the algorithm. More importantly, the algorithm covers only point evidence — observations of the value of a variable at a point in time. As discussed above, in many applications, we observe a variable for an interval, or even for its entire trajectory.

In this paper, we describe an approximate inference algorithm for CTBNs that allows both point and interval evidence. The algorithm uses message passing in a cluster graph, where the clusters do not contain distributions over the cluster variables at individual time points, but distributions over trajectories of the variables through a duration. We provide a new formulation of CTBN parameterization that allows the factors in the clusters to be divided as well as multiplied. With this basic building block, we can execute multiply-marginalize-divide message passing in a cluster graph, as proposed by Lauritzen and Spiegelhalter (1988).

In particular, we can provide an expectation propagation algorithm for CTBNs (Minka, 2001), and prove a characterization of its convergence points as fixed points of a free energy function.

A key benefit of our algorithm is that time is not discretized as part of the model. Rather, the inference algorithm reasons about entire trajectories over the variables in each cluster. Thus, we are not forced to use a fixed, global time granularity for all variables at all times. Rather, the inference algorithm dynamically determines the appropriate granularity to use in order to reason about different clusters in the cluster graph, adapting it to the rate at which the cluster evolves, in the current state of the system. In other words, our inference granularity varies both over variables and over time. This flexiblity allows us to avoid making unnecessary update steps, resulting in possibly significant computational savings over a time-slice approach.

## 2 Continuous Time Bayesian Networks

We begin by briefly reviewing the key definitions of Markov processes and continuous time Bayesian networks, as presented in (Nodelman et al. 2002; 2003).

### 2.1 Representation and Parameterization

A finite state, continuous time, homogeneous Markov process $X_t$ with state space $Val(X) = \{x_1, \ldots, x_n\}$ is described by an initial distribution $P_X^0$ and an $n \times n$ matrix of transition *intensities*:

$$\mathbf{Q}_X = \begin{bmatrix} -q_{x_1} & q_{x_1 x_2} & \cdots & q_{x_1 x_n} \\ q_{x_2 x_1} & -q_{x_2} & \cdots & q_{x_2 x_n} \\ \vdots & \vdots & \ddots & \vdots \\ q_{x_n x_1} & q_{x_n x_2} & \cdots & -q_{x_n} \end{bmatrix},$$

where $q_{x_i x_j}$ is the intensity of transitioning from state $x_i$ to state $x_j$ and $q_{x_i} = \sum_{j \neq i} q_{x_i x_j}$.

Given $\mathbf{Q}_X$ we can describe the transient behavior of $X_t$. If $X_0 = x$ then it stays in state $x$ for an amount of time exponentially distributed with parameter $q_x$. Thus, the probability density function $f$ for $X_t$ remaining at $x$ is $f(q_x, t) = q_x \exp(-q_x t)$ for $t \geq 0$, and the corresponding probability distribution function $F$ for $X_t$ remaining at $x$ for an amount of time $\leq t$ is given by $F(q_x, t) = 1 - \exp(-q_x t)$. The expected time of transitioning is $1/q_x$. Upon transitioning, $X$ shifts to state $x'$ with probability $\theta_{xx'} = q_{xx'}/q_x$. We can view the distribution in terms of the exponential distribution parameter $q_x$, encoding *when* the next transition occurs, and the multinomial parameters $\theta_{xx'}$ ($x \neq x'$), encoding *where* the state transitions.

The distribution over the state of the process $X$ at some future time $t$, $P_X(t)$, can be computed directly from $\mathbf{Q}_X$. If $P_X^0$ is the distribution over $X$ at time 0, then

$$P_X(t) = P_X^0 \exp(\mathbf{Q}_X \cdot t) , \qquad (1)$$

where exp is matrix exponentiation.

NSK extend this framework to encode the joint dynamics of several local variables. In a *continuous time Bayesian network*, each variable $X$ is a Markov process whose parameterization depends on a subset of other variables $\mathbf{U}$.

**Definition 2.1** *A conditional Markov process $X$ is an inhomogeneous Markov process whose intensity matrix varies as a function of the current values of a set of discrete conditioning variables $\mathbf{U}$. It is parameterized using a* conditional intensity matrix (CIM) — $\mathbf{Q}_{X|\mathbf{U}}$ — *a set of homogeneous intensity matrices $\mathbf{Q}_{X|\mathbf{u}}$, one for each instantiation of values $\mathbf{u}$ to $\mathbf{U}$.* ∎

The parameters of $\mathbf{Q}_{X|\mathbf{U}}$ are $\boldsymbol{q}_{X|\mathbf{u}} = \{q_{x|\mathbf{u}} : x \in \mathit{Val}(X)\}$ and $\boldsymbol{\theta}_{X|\mathbf{u}} = \{\theta_{xx'|\mathbf{u}} : x, x' \in \mathit{Val}(X), x \neq x'\}$.

**Definition 2.2** *A continuous time Bayesian network $\mathcal{N}$ over $\boldsymbol{X}$ consists of two components: an* initial distribution $P_{\boldsymbol{X}}^0$, *specified as a Bayesian network $\mathcal{B}$ over $\boldsymbol{X}$, and a* continuous transition model, *specified using a directed (possibly cyclic) graph $\mathcal{G}$ whose nodes are $X \in \boldsymbol{X}$; $\mathbf{U}_X$ denotes the parents of $X$ in $\mathcal{G}$. Each variable $X \in \boldsymbol{X}$ is associated with a conditional intensity matrix, $\mathbf{Q}_{X|\mathbf{U}_X}$.* ∎

### 2.2 Semantics

There are several equivalent ways to define the semantics of a CTBN. NSK define one possible semantics using a "multiplication" operation called *amalgamation* on CIMs. This operation combines two CIMs to produce a single, larger CIM. Amalgamation takes two conditional intensity matrices $\mathbf{Q}_{\boldsymbol{S}_1|\boldsymbol{C}_1}, \mathbf{Q}_{\boldsymbol{S}_2|\boldsymbol{C}_2}$ and combines them to form a new product CIM, $\mathbf{Q}_{\boldsymbol{S}|\boldsymbol{C}} = \mathbf{Q}_{\boldsymbol{S}_1|\boldsymbol{C}_1} * \mathbf{Q}_{\boldsymbol{S}_2|\boldsymbol{C}_2}$ where $\boldsymbol{S} = \boldsymbol{S}_1 \cup \boldsymbol{S}_2$ and $\boldsymbol{C} = (\boldsymbol{C}_1 \cup \boldsymbol{C}_2) - \boldsymbol{S}$. The new CIM contains the intensities for the variables in $\boldsymbol{S}$ conditioned on those of $\boldsymbol{C}$. A basic assumption is that, as time is continuous, variables cannot transition at the same instant. Thus, all intensities corresponding to two simultaneous changes are zero. If the changing variable is from $\boldsymbol{S}_1$, we can look up the correct intensity from the factor $\mathbf{Q}_{\boldsymbol{S}_1|\boldsymbol{C}_1}$. Similarly, if it is from $\boldsymbol{S}_2$, we can look up the intensity from the factor $\mathbf{Q}_{\boldsymbol{S}_2|\boldsymbol{C}_2}$. Intensities along the main diagonal are computed at the end in order to make the rows sum to zero for each instantiation of values to $\boldsymbol{C}$.

**Example 2.3** *Consider a CTBN $A \to B$ with CIMs*

$$\begin{array}{ccc} \mathbf{Q}_A & \mathbf{Q}_{B|a_1} & \mathbf{Q}_{B|a_2} \\ \begin{bmatrix} -1 & 1 \\ 2 & -2 \end{bmatrix} & \begin{bmatrix} -5 & 2 & 3 \\ 2 & -6 & 4 \\ 2 & 5 & -7 \end{bmatrix} & \begin{bmatrix} -7 & 3 & 4 \\ 3 & -8 & 5 \\ 3 & 6 & -9 \end{bmatrix} \end{array}.$$

*The amalgamated CIM assigns intensity 0 to transitions that change both $A$ and $B$ simultaneously. For transitions involving only one of the variables, it simply uses the entry from the appropriate intensity matrix. The resulting matrix, if entries are ordered $(a_1, b_1)$, $(a_2, b_1)$, $(a_1, b_2)$, $(a_2, b_2)$,*

$(a_1, b_3)$, $(a_2, b_3)$, is

$$\mathbf{Q}_{AB} = \begin{bmatrix} -6 & 1 & 2 & 0 & 3 & 0 \\ 2 & -9 & 0 & 3 & 0 & 4 \\ 2 & 0 & -7 & 1 & 4 & 0 \\ 0 & 3 & 2 & -10 & 0 & 5 \\ 2 & 0 & 5 & 0 & -8 & 1 \\ 0 & 3 & 0 & 6 & 2 & -11 \end{bmatrix}.$$

*For example, the entry $(1, 3)$, representing the intensity of transitioning from $(a_1, b_1)$ to $(a_1, b_2)$ is 2, taken from the $(1, 2)$ entry of the matrix $\mathbf{Q}_{B|a_1}$.*

### 2.3 Sufficient Statistics

A CTBN $\mathcal{N}$ defines a probability density over complete trajectories $\sigma$ of the set of variables $\mathbf{X}$. A complete trajectory can be specified as a sequence of states $\mathbf{x}_i$ of $\mathbf{X}$, each with an associated duration. This means we observe every transition of the system from one state to the next and the time at which it occurs. We can conveniently write the density over complete trajectories in terms of the sufficient statistics characterizing the trajectory: $T[x|\mathbf{u}]$ — the amount of time that $X = x$ while $\mathbf{U}_X = \mathbf{u}$; and $M[x, x'|\mathbf{u}]$ — the number of times that $X$ transitions from $x$ to $x'$ while $\mathbf{U}_X = \mathbf{u}$ (Nodelman et al., 2003). If we let $M[x|\mathbf{u}] = \sum_{x'} M[x, x'|\mathbf{u}]$, we can write the density as

$$P_\mathcal{N}(\sigma) = \prod_{X \in \mathbf{X}} L_X(T[X|\mathbf{U}], M[X|\mathbf{U}])$$

where

$$L_X(T[X|\mathbf{U}], M[X|\mathbf{U}]) = \quad (2)$$
$$\prod_{\mathbf{u}} \prod_x (q_{x|\mathbf{u}}^{M[x|\mathbf{u}]} \exp(-q_{x|\mathbf{u}} T[x|\mathbf{u}]) \times \prod_{x' \neq x} \theta_{xx'|\mathbf{u}}^{M[x,x'|\mathbf{u}]})$$

is $X$'s *likelihood contribution* to the overall probability of the trajectory.

## 3 Algorithm Overview

The inference task on which we focus is that of answering probability queries given some partial observations about the current trajectory. Such observations fall into two main types: point evidence and continuous evidence. Point evidence is an observation of the value of a variable at a particular instant in time. Continuous evidence provides the value of a variable throughout an entire interval, which we take take to be a half-closed interval $[t_1, t_2)$. The endpoints of an interval at which a variable is observed do not necessarily correspond to transition points of the variable. They can start at an arbitrary time, contain zero or more transitions, and end at an arbitrary time.

Without loss of generality, we can partition our evidence into a sequence of intervals of constant continuous evidence, possibly punctuated by point evidence or observed transitions. Within each interval, the set of variables we observe and their values are both constant. Note that constant continuous evidence includes the possibility of no evidence on the interval. This creates a sequence of distinguished time points $t_1, \ldots, t_n$ with constant continuous evidence $\mathbf{e}_i^s$ on every interval $[t_i, t_{i+1})$ and possible point evidence or observed transition $\mathbf{e}_i^p$ at each $t_i$. Both $\mathbf{e}_i^s$ and $\mathbf{e}_i^p$ are assignments to some subset of the variables in $\mathbf{X}$.

**Example 3.1** *Consider a system over the variables $X, Y, Z$. One set of evidence might have: $X = x_1, Y = y_1$ for the interval $[0, 0.7)$; $Y = y_2, Z = z_1$ for the interval $[0.7, 1.1)$, $Z = z_2$ at $t = 1.1$; $X = x_1$ for the interval $[1.1, 2)$; and $Y = y_1$ at $t = 1.5$. The distinguished time points are $0, 0.7, 1.1, 1.5, 2$. Note that the value of $X$ may have changed 0 or more times in the interval $[0.7, 1.1)$. The value of $Y$ changed from $y_1$ to $y_2$ at exactly $0.7$. We observe a transition for $Z$ at $t = 1.1$, and an isolated observation of $Y$'s value at $t = 1.5$.*

As NSK discuss, there is a range of query types that can be answered using a CTBN. These include the value of a variable at a given time, but also the time at which a variable first takes a particular value, or the expected number of times that a variable changes value. We propose an algorithm that can address all of these types of query, given both point and continuous evidence.

Our algorithm uses message passing in cluster graphs, of which clique tree algorithms are a special case. In cluster graph algorithms, we construct a graph whose nodes correspond to clusters of variables, and pass messages between these clusters to produce an alternative parameterization, in which the marginal distribution of the variables in each cluster can be read directly from the cluster. In discrete graphical models, when the cluster graph is a clique tree, two passes of message passing produce exact marginals. In *generalized belief propagation* (Yedidia et al., 2000), message passing is applied to a graph which is not a clique tree, in which case the algorithm may not converge, and produces only approximate solutions. There are several forms of message passing algorithm. Our algorithm is based on multiply-marginalize-divide scheme of Lauritzen and Spiegelhalter (1988), which we now briefly review.

At a high level, a cluster graph is defined in terms of a set of clusters $\mathcal{C}_i$, whose *scope* is some subset of the variables $\mathbf{X}$. Clusters are connected to each other by edges, along which messages are passed. The edges are annotated with a set of variables called a *sepset* $\mathcal{S}_{i,j}$ which is the set of variables in $\mathcal{C}_i \cap \mathcal{C}_j$. The messages passed over an edge $\mathcal{C}_i$—$\mathcal{C}_j$ are *factors* over the scope $\mathcal{S}_{i,j}$.

Each cluster $\mathcal{C}_i$ maintains a *potential* $\pi_i$, a factor which reflects its current beliefs over the variables in its scope. Each edge similarly maintains a message $\mu_{i,j}$, which encodes the last message sent over the edge. The potentials are initialized with a product of some subset of factors parameterizing the model (CIMs in our setting). Messages are initialized to be uninformative. Clusters then send messages to each other, and use incoming messages to update their beliefs over the variables in their scope. The message $\delta_{i \to j}$ from $\mathcal{C}_i$ to $\mathcal{C}_j$ is the marginal distribution $\mathcal{S}_{i,j}$ according to $\pi_i$. The neighboring cluster $\mathcal{C}_j$ assimilates this message by multiplying it into $\pi_i$, but avoids double-counting by first dividing by the stored message $\mu_{i,j}$. Thus, the message update takes the form $\pi_j \leftarrow \pi_j \cdot \frac{\delta_{i \to j}}{\mu_{i,j}}$.

In our algorithm, the clusters do not represent factors over values of random variables. Rather, cluster potentials and messages both encode measures over entire trajectories of the variables in their scope.

**Example 3.2** *Consider a CTBN $A \to B \to C \to D$. We can form a clique tree $\{A, B\}$—$\{B, C\}$—$\{C, D\}$, where the $\{A, B\}$ cluster, for example, contains the CIMs $\mathbf{Q}_A, \mathbf{Q}_{B|A}$. Note that the message from this cluster to the $\{B, C\}$ cluster is a marginal encoding a distribution over $B$'s trajectories. Although the joint $A, B$ distribution is a homogeneous Markov process over $A, B$, the marginal distribution over $B$ is not typically a homogeneous Markov process.*

As in this example, the marginal distributions that form the messages are not homogeneous Markov processes; indeed, the exact marginal distributions for the true joint distribution can be arbitrarily complex, requiring a number of parameters which grows exponentially with the size of the network. Thus, we cannot pass messages exactly without giving up the computational efficiency of the algorithm. We address this issue using the *expectation propagation (EP)* approach (Minka, 2001), which performs approximate message passing in cluster graphs.

EP addresses the problem where messages can be too complex to represent and manipulate by using approximate messages, projecting each message $\delta_{i \to j}$ into a compactly representable space so as to minimize the KL-divergence between $\delta_{i \to j}$ and its approximation $\hat{\delta}_{i \to j}$. In a prototypical example (Minka, 2001), the cluster potentials and therefore the sepset marginals are mixtures of Gaussians, which are projected into the space of Gaussian distributions in the message approximation step. For messages in the exponential family, $\arg\min_{\hat{\delta}_{i \to j}} \mathbf{D}(\delta_{i \to j} || \hat{\delta}_{i \to j})$ can be obtained by matching moments of the distribution. EP can be applied to clique trees or to general cluster graphs. Note that, even in clique trees, the algorithm does not generally converge after two passes of message passing (as it does in exact inference), so that multiple iterations are generally required, and convergence is not guaranteed.

In our application of EP, we use conditional intensity matrices (CIMs), reduced to match the evidence, to encode the cluster potentials; we approximate the messages in the cluster graph as homogeneous Markov processes, using a KL-divergence projection. To apply the EP algorithm to clusters of this form, we need to define basic operations over CIMs. First, we need to define the operations of multiplying and dividing CIMs, used in the message update step. Second, we need to describe the construction of initial potentials from CIMs, and how they account for the evidence. Finally, we need to show how to perform approximate marginalization of CIMs, used to compute the message the approximate marginals of a cluster potential over its sepset. We begin by describing these operations in the next section, and then present the algorithm in its entirety in Sec. 5.

We note that a similar approach — of encoding clusters as CIMs and approximating messages as homogeneous Markov processes — was used in the original clique tree algorithm of Nodelman et al. (2002), but with important differences. Most importantly, the new operations on CIMs allow us to to deal with continuous evidence rather than just point evidence. Second, the NSK algorithm was based on multiply-marginalize message passing scheme of Shafer and Shenoy (1990) algorithm, whereas our algorithm is based on multiply-marginalize-divide scheme of Lauritzen and Spiegelhalter (1988). Second, our algorithm performs approximate marginalization so as to minimize KL-divergence, a more principled approach. As a consequence, we can use the iterative EP algorithm for message propagation, improving the quality of approximation. As an instance of EP, our algorithm has the property that it converges to fixed points of the approximate free energy function, subject to calibration constraints on the approximate messages. Finally, and It also allows us to

## 4 Basic Operations

The basis for our algorithm is a reformulation of CIMs that supports the key operations required for message passing in EP: CIM product and division, incorporating evidence into a CIM, and approximate CIM marginalization.

### 4.1 Amalgamating CIMs

A CIM $\mathbf{Q}_{S|C}$ over variables $S \subseteq X$ conditioned on $C \subset X$ defines the dynamics of $S$ given $C$. We can rewrite $\mathbf{Q}_{S|C}$ as a single block matrix over the joint space $S \times C$:

$$\mathbf{Q}_{S|C} = \left[ \begin{array}{cccc} \mathbf{Q}_{S|c_1} & 0 & \cdots & 0 \\ 0 & \mathbf{Q}_{S|c_2} & \cdots & 0 \\ \vdots & \vdots & \ddots & \vdots \\ 0 & 0 & \cdots & \mathbf{Q}_{S|c_N} \end{array} \right].$$

The CIM $\mathbf{Q}_{S|C}$ induces a distribution $\phi(S|C)$ over the dynamics of $S$ given $C$. Analogous to Eq. (1), exponentiating the CIM by taking $\phi(S|C)^t = \exp(\mathbf{Q}_{S|C} \cdot t)$ gives u the probability that, if we start with $S = s$ and continue for $t$ time, we end up at $S = s'$, given that $C = c$ for the entire time period. Thus, we can view a CIM as the logarithm of the distribution over the (conditional) system dynamics.

We can now redefine the amalgamation operation in terms of this representation of CIMs. First, note that if we have a CIM $\mathbf{Q}_{S'|C'}$ where $S' \subseteq S$ and $C' \subseteq C$, we can embed it within a matrix over $S \times C$ by embedding multiple copies of $\mathbf{Q}_{S'|C'}$ in the new, larger matrix. The resulting matrix would look just as above, except with repeated copies of $\mathbf{Q}_{S'|C'}$. We can choose the order of the states in the matrix arbitrarily.

**Definition 4.1** Amalgamation *is an operation which takes two CIMS $\mathbf{Q}_{S_1|C_1}$, $\mathbf{Q}_{S_2|C_2}$, and forms the new CIM $\mathbf{Q}_{S|C}$ where $S = S_1 \cup S_2$ and $C = (C_1 \cup C_2) \setminus S$. First we expand $\mathbf{Q}_{S_1|C_1}$ and $\mathbf{Q}_{S_2|C_2}$ into single matrices over $S \times C$ and then define the amalgamated matrix as the sum $\mathbf{Q}_{S|C} = \mathbf{Q}_{S_1|C_1} + \mathbf{Q}_{S_2|C_2}$. The inverse of amalgamation is computed by matrix subtraction.* ∎

**Example 4.2** *Consider the CTBN from Example 2.3. We expand each of $\mathbf{Q}_A$ and $\mathbf{Q}_{B|A}$ into a single matrix over the space $A \times B$, in the order $(a_1, b_1)$, $(a_2, b_1)$, $(a_1, b_2)$, $(a_2, b_2)$, $(a_1, b_3)$, $(a_2, b_3)$, obtaining:*

$$\mathbf{Q}_A = \begin{bmatrix} -1 & 1 & 0 & 0 & 0 & 0 \\ 2 & -2 & 0 & 0 & 0 & 0 \\ 0 & 0 & -1 & 1 & 0 & 0 \\ 0 & 0 & 2 & -2 & 0 & 0 \\ 0 & 0 & 0 & 0 & -1 & 1 \\ 0 & 0 & 0 & 0 & 2 & -2 \end{bmatrix}$$

$$\mathbf{Q}_{B|A} = \begin{bmatrix} -5 & 0 & 2 & 0 & 3 & 0 \\ 0 & -7 & 0 & 3 & 0 & 4 \\ 2 & 0 & -6 & 0 & 4 & 0 \\ 0 & 3 & 0 & -8 & 0 & 5 \\ 2 & 0 & 5 & 0 & -7 & 0 \\ 0 & 3 & 0 & 6 & 0 & -9 \end{bmatrix}.$$

*The amalgamation of these two CIMs is given by the matrix addition $\mathbf{Q}_A + \mathbf{Q}_{B|A}$, producing precisely the matrix $\mathbf{Q}_{AB}$ shown in Example 2.3.*

The use of addition for amalgamation of CIMs is very natural when we consider its interpretation as the logarithm of the system dynamics. Specifically, adding CIMs is equivalent to multiplying their distributions. Recalling that CIMs directly parametrize the instantaneous behavior and examining the distribution as $t \to 0$,

$$\begin{aligned}
\phi(S|C)^t &= \phi(S_1|C_1)^t \cdot \phi(S_2|C_2)^t \\
&= \exp(\mathbf{Q}_{S_1|C_1} \cdot t) \cdot \exp(\mathbf{Q}_{S_2|C_2} \cdot t) \\
&= (\mathbf{I} + \mathbf{Q}_{S_1|C_1} t + O(t^2)) \cdot (\mathbf{I} + \mathbf{Q}_{S_2|C_2} t + O(t^2)) \\
&= \mathbf{I} + (\mathbf{Q}_{S_1|C_1} + \mathbf{Q}_{S_2|C_2})t + O(t^2) \\
&= \exp((\mathbf{Q}_{(S_1|C_1} + \mathbf{Q}_{S_2|C_2})t) \\
&= \exp(\mathbf{Q}_{S|C} \cdot t) \ .
\end{aligned}$$

If we amalgamate all the CIMs of a CTBN $\mathcal{N}$, we get a single intensity matrix encoding the distribution over the dynamics of the entire system:

$$P_\mathcal{N} = \prod_{X \in \mathbf{X}} \phi(X|\mathbf{U}_X) = \exp\left(\sum_{X \in \mathbf{X}} \mathbf{Q}_{X|\mathbf{U}_X}\right).$$

This definition of amalgamation handles not only full CIMs, but also CIMs that are reduced to account for conditioning on continuous evidence, as we discuss next.

## 4.2 Incorporating Evidence into CIMs

Point observations about the system state affect our distribution over the state at a single point in time, which in turn, affects the distribution over the behavior of the system. But it does not affect our distribution over the dynamics as parameterized by the CIMs. By contrast, consider continuous evidence, as in Example 4.2. If we condition on the continuous evidence that $A = a_1$ for all $t \in [0, 1]$, then the dynamics of $Z$ during that interval is described solely by $\mathbf{Q}_{B|a_1}$ rather than a mixture of $\mathbf{Q}_{B|a_1}$ and $\mathbf{Q}_{B|a_2}$. An observation over an interval restricts our transition dynamics to remain within a subset of the full state space for the duration of the interval.

To account for such evidence, we *reduce* the CIM — eliminate the rows and columns of the CIM that correspond to states inconsistent with the evidence. In the special case where we are conditioning $\mathbf{Q}_{S|C}$ on evidence $\mathbf{e}$ over some variable(s) in the conditioning set $C$, the result is a CIM $\mathbf{Q}_{S|C,\mathbf{e}}$ that represents the conditional distribution $\phi(S|C, \mathbf{e})$. More generally, when we have evidence $\mathbf{e}_1$ within $S$ and $\mathbf{e}_2$ within $C$, the reduced CIM represents the unnormalized conditional distribution $\phi(S, \mathbf{e}_1|C, \mathbf{e}_2)$. In this case, the reduced intensity matrix $\mathbf{Q}_{S,\mathbf{e}_1|C,\mathbf{e}_2}$ will have rows that sum to negative numbers. These negative numbers represent "extra" intensity with which we would normally leave the subsystem (if not for the evidence), and represent the probability flowing out of the subsystem. Note that a reduced intensity matrix $\phi(S, \mathbf{e})$ cannot, in general, be normalized and represented as an intensity matrix.

**Example 4.3** *Consider the system over $A$ and $B$ described in Example 4.2. If we want to incorporate the continuous evidence that $B = b_1$ for time $t \in [0, 1]$, we use the reduced intensity matrix*

$$\mathbf{Q}_{A, b_1} = \begin{bmatrix} -6 & 1 \\ 2 & -9 \end{bmatrix} \ .$$

*As described above, the rows sum to negative numbers, whose magnitude corresponds to the intensity with which we would normally leave the subsystem when $B = b_1$.*

## 4.3 Marginalizing CIMs

Clusters in our cluster tree are associated with unnormalized CIMs, perhaps reduced by the incorporation of continuous evidence. In most cases, the marginal dynamics of such a CIM over a subset of variables cannot be described using an unconditional intensity matrix. Indeed, in general, the marginal distribution over a single variable $X$ can only be correctly described by constructing the entire joint intensity matrix, and considering its marginal distribution over $X$. However, we can approximately marginalize factors — products of (reduced) CIMs — by projecting them into the space of distributions represented as unconditional intensity matrices.

More precisely, consider the distribution $\phi(\boldsymbol{S}) \propto P_{\boldsymbol{S}}^0 \exp(\mathbf{Q}_{\boldsymbol{S}} t)$ described by a (possibly reduced) intensity matrix $\mathbf{Q}_{\boldsymbol{S}}$. This distribution induces a marginal distribution $\phi(\boldsymbol{V})$ over the dynamics of $\boldsymbol{V}$ for any subset $\boldsymbol{V} \subset \boldsymbol{S}$. We would like to project $\phi(\boldsymbol{V})$ onto the space of distributions representable by the intensity matrix $\hat{\mathbf{Q}}_{\boldsymbol{V}}$, by minimizing the Küllback-Leibler divergence; specifically, we want to compute $\arg\min_{\hat{P}_{\boldsymbol{V}}} \mathbf{D}(P_{\boldsymbol{V}} \| \hat{P}_{\boldsymbol{V}})$ where $\hat{P}_{\boldsymbol{V}}(t) = P_{\boldsymbol{V}}^0 \exp(\hat{\mathbf{Q}}_{\boldsymbol{V}} t)$. As the set of distributions representable by an intensity matrix is in the exponential family, we can minimize the KL-divergence over an interval $[t_1, t_2]$ by choosing $\hat{P}_{\boldsymbol{V}}(t)$ to match the moments of $P_{\boldsymbol{V}}(t)$ over $[t_1, t_2]$.

Importantly, a CIM in isolation, or even an unreduced intensity matrix, does not define a distribution over trajectories. To define a distribution and the requisite moments, we need an initial state distribution $P_{\boldsymbol{V}}^0$ at time $t_1$ and the duration of the interval $[t_1, t_2]$. Given a reduced CIM $\phi(\boldsymbol{V}, \mathbf{e})$ over the interval $[t_1, t_2]$, we can obtain the conditional distribution over the system behavior by normalizing the distribution: $\phi(\boldsymbol{V} | \mathbf{e}) = \frac{1}{Z} \exp(\mathbf{Q}_{\boldsymbol{V},\mathbf{e}} \cdot (t_2 - t_1))$, where $Z$ is the *partition function* representing the probability of the evidence: $Z = \int_{t_1}^{t_2} P_{\boldsymbol{V}}^0 \exp(\mathbf{Q}_{\boldsymbol{V},\mathbf{e}} \cdot t) dt$. Note that $Z$ is a function of the amount of time the evidence persists and of the distribution $P_{\boldsymbol{V}}^0$ over the state at the beginning of the evidence.

To match moments, we must compute the expected sufficient statistics over the interval $[t_1, t_2]$ for the variables in $\boldsymbol{S}$. These expected sufficient statistics are $\mathbf{E}[T[j]]$, the expected amount of time in each state $j$, and $\mathbf{E}[M[j,k]]$, the expected number of transitions from $j$ to $k$. For simplicity, assume that the evidence is constant throughout the interval. We can compute sufficient statistics for the more general case using a forward-backward algorithm (see Nodelman et al. (2005) for details and derivations). Let $\boldsymbol{\Delta}_{j,k}$ be a matrix with a one in row $j$, column $k$, and zeros everywhere else. Let $\mathbf{e}$ be a column vector of ones. Then for each instantiation $j$ of $\boldsymbol{S}$, we compute $\mathbf{E}[T[j]]$ as

$$c \int_{t_1}^{t_2} P_{\boldsymbol{S}}^0 \exp(\mathbf{Q}_{\boldsymbol{S}}(t - t_1)) \boldsymbol{\Delta}_{j,j} \exp(\mathbf{Q}_{\boldsymbol{S}}(t_2 - t)) \mathbf{e} \, dt \, ;$$

that is, we integrate over the probability of remaining in state $j$. The normalization constant $c$ makes the expected amount of time over all states sum to $t_2 - t_1$. Similarly, for each pair of instantiations $j, k$, we compute $\mathbf{E}[M[j,k]]$ as

$$c \, q_{jk} \int_{t_1}^{t_2} P_{\boldsymbol{S}}^0 \exp(\mathbf{Q}_{\boldsymbol{S}}(t - t_1)) \boldsymbol{\Delta}_{j,k} \exp(\mathbf{Q}_{\boldsymbol{S}}(t_2 - t)) \mathbf{e} \, dt \, ;$$

that is, we integrate over the instantaneous probability of transitioning and use the same normalization constant. These integrals are guaranteed to be finite for any finite interval $[t_1, t_2]$.

We can calculate the set of these integrals for all $j$ and $k$ simultaneously (as a set of differential equations) via the Runge-Kutta method of fourth order with adaptive step size. This method traverses the interval in small discrete steps each of which has a constant number of matrix multiplications. Thus, the main factor in the complexity of this algorithm is the number of steps which is a function of the step size.

Importantly, the step size is *adaptive* and not fixed. The intensities of the $\mathbf{Q}_S$ matrix represent rates of evolution for the variables in the cluster, so larger intensities mean a faster rate of change which usually requires a smaller step size. We begin by setting the step size proportional to the inverse of the largest intensity in $\mathbf{Q}_S$. The step size thus varies across different clusters and is sensitive to the current evidence. Also, following Press et al. (1992), we use a standard adaptive procedure that allows larger steps to be taken when possible based on error estimates.

Given the expected sufficient statistics over $\boldsymbol{S}$, we can calculate $\mathbf{E}[T[\boldsymbol{v}]]$, the expected amount of time in each instantiation $\boldsymbol{v}$ of $\boldsymbol{V}$, and $\mathbf{E}[M[\boldsymbol{v}, \boldsymbol{v}']]$, the expected number of transitions from $\boldsymbol{v}$ to $\boldsymbol{v}'$. We also compute the total number of expected transitions from $\boldsymbol{v}$, $\mathbf{E}[M[\boldsymbol{v}]] = \sum_{\boldsymbol{v}'} M[\boldsymbol{v}, \boldsymbol{v}']$. We can now match moments, setting the parameters of $\hat{\mathbf{Q}}_{\boldsymbol{V}}$ to be the maximum likelihood parameters (Nodelman et al., 2003),

$$q_{\boldsymbol{v}} = \frac{\mathbf{E}[M[\boldsymbol{v}]]}{\mathbf{E}[T[\boldsymbol{v}]]}; \quad \theta_{\boldsymbol{v}\boldsymbol{v}'} = \frac{\mathbf{E}[M[\boldsymbol{v},\boldsymbol{v}']]}{\mathbf{E}[M[\boldsymbol{v}]]} \, . \tag{3}$$

We write $\hat{\phi}(\boldsymbol{V}) = \mathrm{marg}_{\boldsymbol{S} \setminus \boldsymbol{V}}^{P^0, T}(\phi(\boldsymbol{S}))$ for the distribution parameterized by $\hat{\mathbf{Q}}_{\boldsymbol{V}}$.

**Example 4.4** *Consider the system over A and B described in Example 4.2. If we assume a uniform initial distribution and that we want to use this approximation for unit time ($T = 1$), then the matrix of expected sufficient statistics*

$$\bar{M}[(a,b),(a',b')] = \begin{bmatrix} - & .18 & .36 & 0 & .54 & 0 \\ .24 & - & 0 & .35 & 0 & .47 \\ .45 & 0 & - & .23 & .91 & 0 \\ 0 & .42 & .28 & - & 0 & .70 \\ .41 & 0 & 1.03 & 0 & - & .21 \\ 0 & .39 & 0 & .78 & .26 & - \end{bmatrix},$$

*and* $\bar{T}[(a,b)] = \begin{bmatrix} .18 & .12 & .23 & .14 & .21 & .13 \end{bmatrix}$. *Combining sufficient statistics for $b_1$ (rows 1,2), $b_2$ (rows 3,4), and $b_3$ (rows 5,6) we get the following matrix of expected sufficient statistics over B*

$$\bar{M}[b,b'] = \begin{bmatrix} - & .71 & 1.01 \\ .87 & - & 1.61 \\ .80 & 1.81 & - \end{bmatrix}.$$

*and* $\bar{T}[b] = \begin{bmatrix} .30 & .37 & .33 \end{bmatrix}$. *With the expected sufficient statistics over B, we can compute the parameters of $\hat{\phi}(B) = \mathrm{marg}_A^{P^0,1}(\phi(A,B))$,*

$$\hat{\mathbf{Q}}_B = \begin{bmatrix} -5.73 & 2.37 & 3.36 \\ 2.35 & -6.70 & 4.35 \\ 2.42 & 5.49 & -7.91 \end{bmatrix}.$$

There is an additional subtlety to the computation if $\mathbf{Q}_S$ is conditioned on continuous evidence and has negative row sums (representing the probability of the evidence as discussed in Sec. 4.2). In this case, we must account for the extra intensity of leaving the subsystem entirely when computing the expected number of transitions out of each state $\mathbf{E}[M[\boldsymbol{v}]]$. To do so, we add an extra state $\iota$ to $\mathbf{Q}_S$ before computing the expected sufficient statistics. For each instantiation $\boldsymbol{s}$ of $\boldsymbol{S}$, the intensity of entering the extra state — $q_{\boldsymbol{v}\iota}$ — makes the row sum to zero. Then, when we compute $\mathbf{E}[M[\boldsymbol{v}]]$, we also include $\mathbf{E}[M[\boldsymbol{v}, \iota]]$ the expected number of transitions to the extra state, and use Eq. (3). In the computation, the normalization constant $c$ makes the total time spent in all states except $\iota$ sum to $t_2 - t_1$. Note that, as $\iota$ does not correspond to any instantiation $\boldsymbol{v}$, we have that $\sum_{\boldsymbol{v}'} \theta_{\boldsymbol{v}\boldsymbol{v}'} < 1$, and therefore the row sums in the resulting intensity matrix will also be negative. This corresponds to the fact that our marginalized intensity matrix approximates the marginal of $P(\mathbf{e}, \boldsymbol{S} \mid \boldsymbol{C})$ in this case.

**Example 4.5** *Continuing Example 4.3, we add a new state $\iota$, resulting in a new CIM:*

$$\mathbf{Q}_{A,b_1} = \begin{bmatrix} -6 & 1 & 5 \\ 2 & -9 & 7 \\ 0 & 0 & 0 \end{bmatrix},$$

*where the last row/column correspond to the absorbing state $\iota$. Assume a uniform initial distribution over the states of $A$ and that we are in this subsystem for total time $T = 1$. Then, calculating the integrals by Runge-Kutta without normalizing yields the unnormalized matrix over transitions of $A$ including the additional state $\iota$,*

$$\begin{bmatrix} - & .105 & .526 \\ 0.134 & - & .470 \\ 0 & 0 & - \end{bmatrix},$$

*and an unnormalized vector over the amount of time in each state,* $\begin{bmatrix} .105 & .067 & .828 \end{bmatrix}$. *The normalization constant $c = 1/(.105 + .067) = 5.81$. So the expected sufficient statistics (given that we spend no time in $\iota$) are*

$$\bar{M}[a, a'] = \begin{bmatrix} - & 0.61 & 3.05 \\ 0.78 & - & 2.73 \\ 0 & 0 & - \end{bmatrix},$$

*and $\bar{T}[a] = \begin{bmatrix} .61 & .39 & 0 \end{bmatrix}$. When we compute parameters with these statistics, we find that we get back the same $\mathbf{Q}_{A,b_1}$ as above because we have not incorporated any additional evidence. Incorporating evidence will generally lead to a different intensity matrix, as in Example 5.1.*

## 5 Expectation Propagation

Based on these operations, we can describe a new message propagation algorithm for CTBNs. As discussed above, unlike the algorithm of Nodelman et al. (2002), the new algorithm uses product-marginalize-divide message passing, as in the clique tree algorithm of Lauritzen and Spiegelhalter (1988). As a consequence, when using approximate projection, we can apply the algorithm iteratively, as in expectation propagation, with the goal of improving our estimates.

### 5.1 EP for Segments

We first consider the message propagation algorithm for one segment of our trajectory, with constant continuous evidence. The generalization to multiple segments follows.

We first construct the cluster tree for the graph $\mathcal{G}$. This procedure is exactly the same as in Bayesian networks — cycles do not introduce new issues. We simply moralize the graph, connecting all parents of a node with undirected edges, and then make all the remaining edges undirected. If we have a cycle, it simply turns into a loop in the resulting undirected graph. We then select a set of clusters $\mathcal{C}_i$. These clusters can be selected so as to produce a clique tree for the graph, using any standard method for constructing such trees. Or, we can construct a loopy cluster graph, and use generalized belief propagation. The message passing scheme is the same in both cases.

Let $\boldsymbol{A}_i \subseteq \mathcal{C}_i$ be the set of variables whose factors we associate with cluster $\mathcal{C}_i$. Let $N_i$ be the set of neighboring clusters for $\mathcal{C}_i$ and let $\mathcal{S}_{i,j}$ be the set of variables in $\mathcal{C}_i \cap \mathcal{C}_j$. We also compute, for each cluster $\mathcal{C}_i$, the initial distribution $P^0_{\mathcal{C}_i}$ using standard BN inference on the network $\mathcal{B}$. After initialization, the algorithm is

**Procedure** *CTBN-Segment-EP*$(P^0, T, \mathbf{e}, \mathcal{G})$
   1. For each cluster $\mathcal{C}_i$
      $\pi_i \leftarrow \prod_{X \in \boldsymbol{A}_i} \phi(X, \mathbf{U}_X, \mathbf{e})$
   2. For each edge $\mathcal{C}_i$—$\mathcal{C}_j$
      $\mu_{i,j} \leftarrow \mathbf{1}$
Loop until convergence:
   3. Choose $\mathcal{C}_i$—$\mathcal{C}_j$
   4. Send-Message$(i, j, P^0_{\mathcal{C}_i}, T)$

**Procedure** *Send-Message*$(i, j, P^0, T)$
   1. $\delta_{i \to j} \leftarrow \text{marg}^{P^0, T}_{\mathcal{C}_i \setminus \mathcal{S}_{i,j}}(\pi_i)$
   2. $\pi_j \leftarrow \pi_j \cdot \frac{\delta_{i \to j}}{\mu_{i,j}}$
   3. $\mu_{i,j} \leftarrow \delta_{i \to j}$

It takes the initial distributions over the clusters $P^0$, an amount of time $T$, and possibly some continuous evidence $\mathbf{e}$ which holds for the total time $T$. We use $\phi(\cdot, \mathbf{e})$ to denote the CIM reduced by continuous evidence $\mathbf{e}$ if applicable. The algorithm iteratively selects an edge $(i, j)$ in the cluster graph, and passes a message from $\mathcal{C}_i$ to $\mathcal{C}_j$. In clique tree propagation, we might select edges so as to iteratively perform an upward and downward pass. In generalized belief propagation, we might use a variety of message passing schemes. Convergence occurs when messages cease to affect the potentials which means that neighboring clusters $\mathcal{C}_i$ and $\mathcal{C}_j$ agree on the approximate marginals over the variables $\mathcal{S}_{i,j}$.

The basic factor operations are performed as described in Sec. 4. Specifically, let $\rho(\cdot)$ be a function taking factors to their CIM parameterization. For the initial potentials, $\rho(\pi_i)$ is computed by adding the intensity matrices $\mathbf{Q}_{X|\mathbf{U}_X}$ reduced by evidence $\mathbf{e}$ for $X \in \mathbf{A}_i$. Also, $\rho(\mathbf{1})$ is an intensity matrix of zeros. Factor product is implemented as addition of intensity matrices, and factor division as subtraction, so that $\rho(\pi_j \cdot \frac{\delta_{i \to j}}{\mu_{i,j}}) = \rho(\pi_j) + \rho(\delta_{i \to j}) - \rho(\mu_{i,j})$. Marginalization is implemented by computing the expected sufficient statistics, using the evidence $\mathbf{e}$, the time period $T$, and the initial distribution $P^0$, as described in Sec. 4.3.

**Example 5.1** *Assume we have a CTBN with 4 binary variables and graph $A \to B \to C \to D$ with CIMs*

$$\begin{array}{ccc} \mathbf{Q}_A & \mathbf{Q}_{B|a_1} & \mathbf{Q}_{B|a_2} \\ \begin{bmatrix} -1 & 1 \\ 1 & -1 \end{bmatrix} & \begin{bmatrix} -1 & 1 \\ 10 & -10 \end{bmatrix} & \begin{bmatrix} -10 & 10 \\ 1 & -1 \end{bmatrix}, \end{array}$$

*where $\mathbf{Q}_{C|B}$ and $\mathbf{Q}_{D|C}$ have the same parameterization as $\mathbf{Q}_{B|A}$. So $A$ switches randomly between states $a_1$ and $a_2$, and each child tries to match the behavior of its parent. Suppose we have a uniform initial distribution over all variables except $D$ which starts in state $d_1$ and remains in that state for unit time (T=1). Our cluster tree is AB—BC—CD and our initial potentials are:*

$$\rho(\pi_1) = \mathbf{Q}_{AB} = \begin{bmatrix} -2 & 1 & 1 & 0 \\ 1 & -11 & 0 & 10 \\ 10 & 0 & -11 & 1 \\ 0 & 1 & 1 & -2 \end{bmatrix},$$

$$\rho(\pi_2) = \mathbf{Q}_{BC} = \begin{bmatrix} -1 & 0 & 1 & 0 \\ 0 & -10 & 0 & 10 \\ 10 & 0 & -10 & 0 \\ 0 & 1 & 0 & -1 \end{bmatrix},$$

$$\rho(\pi_3) = \mathbf{Q}_{Cd_1} = \begin{bmatrix} -1 & 0 \\ 0 & -10 \end{bmatrix}.$$

*Our initial messages are*

$$\rho(\delta_{1 \to 2}) = \begin{bmatrix} -2.62 & 2.62 \\ 2.62 & -2.62 \end{bmatrix} \quad \rho(\delta_{3 \to 2}) = \begin{bmatrix} -1 & 0 \\ 0 & -10 \end{bmatrix}$$

*These messages leave $\pi_1$, $\pi_3$ unchanged and give us:*

$$\rho(\pi_2) = \begin{bmatrix} -4.62 & 2.62 & 1 & 0 \\ 2.62 & -13.62 & 0 & 10 \\ 10 & 0 & -22.62 & 2.62 \\ 0 & 1 & 2.62 & -13.62 \end{bmatrix}.$$

*Our next messages are:*

$$\rho(\delta_{2 \to 1}) = \begin{bmatrix} -5.02 & 2.62 \\ 2.62 & -8.57 \end{bmatrix} \quad \rho(\delta_{2 \to 3}) = \begin{bmatrix} -4.42 & 3.42 \\ 3.62 & -13.62 \end{bmatrix}.$$

*These leave $\pi_2$ unchanged and give us*

$$\rho(\pi_1) = \begin{bmatrix} -4.40 & 1 & 1 & 0 \\ 1 & -13.40 & 0 & 10 \\ 10 & 0 & -16.94 & 1 \\ 0 & 1 & 1 & -7.94 \end{bmatrix},$$

$$\rho(\pi_3) = \begin{bmatrix} -4.42 & 3.42 \\ 3.62 & -13.62 \end{bmatrix}.$$

*Now $\delta_{3 \to 2}$ would have no effect on $\pi_2$, however,*

$$\rho(\delta_{1 \to 2}) = \begin{bmatrix} -5.34 & 2.95 \\ 3.31 & -9.26 \end{bmatrix}$$

*which changes $\pi_2$ so that*

$$\rho(\pi_2) = \begin{bmatrix} -4.95 & 2.95 & 1 & 0 \\ 3.31 & -14.31 & 0 & 10 \\ 10 & 0 & -22.95 & 2.95 \\ 0 & 1 & 3.31 & -14.31 \end{bmatrix}.$$

*Our next messages are*

$$\rho(\delta_{2 \to 1}) = \begin{bmatrix} -5.39 & 2.95 \\ 3.31 & -9.16 \end{bmatrix} \quad \rho(\delta_{2 \to 3}) = \begin{bmatrix} -4.43 & 3.43 \\ 3.76 & -13.76 \end{bmatrix}.$$

*This gives us*

$$\rho(\pi_1) = \begin{bmatrix} -4.45 & 1 & 1 & 0 \\ 1 & -13.45 & 0 & 10 \\ 10 & 0 & -16.85 & 1 \\ 0 & 1 & 1 & -7.85 \end{bmatrix},$$

$$\rho(\pi_3) = \begin{bmatrix} -4.43 & 3.43 \\ 3.76 & -13.76 \end{bmatrix}.$$

*At this point we have converged. If we use $\pi_1$ to compute the distribution over $A$ at time 1, we get $\begin{bmatrix} .703 & .297 \end{bmatrix}$. If we do exact inference by amalgamating all the factors and exponentiating, we get $\begin{bmatrix} .738 & .262 \end{bmatrix}$.*

### 5.2 EP for Trajectories

When we have a trajectory containing multiple segments of continuous evidence, we apply this algorithm separately to every segment, passing information from one to the other in the form of distributions. More precisely, consider a trajectory defining a sequence of time points $t_1, \ldots, t_n$, with constant continuous evidence $\mathbf{e}_i^s$ on every interval $[t_i, t_{i+1})$ and possible point evidence or observed transition $\mathbf{e}_i^p$ at each $t_i$. We construct a sequence of cluster graphs $\mathcal{G}_{t_i, t_{i+1}}$, each over a segment $[t_i, t_{i+1})$. Starting from the initial segment, we run inference on each cluster graph using *CTBN-Segment-EP*, and compute the resulting distribution at time $t_{i+1}$; we condition on any point evidence or the observed transition, and use the new distribution as the initial distribution from the next interval. The formal algorithm is as follows:

**Procedure** *CTBN-Filter-EP*$(P^0, \langle t_0, \ldots, t_n \rangle,$
$\quad \langle \mathbf{e}_1^s, \ldots, \mathbf{e}_n^s \rangle, \langle \mathbf{e}_1^p, \ldots, \mathbf{e}_n^p \rangle)$
For $i = 0, \ldots, n-1$
   1. Construct a cluster graph $\mathcal{G}_{t_i, t_{i+1}}$
   2. *CTBN-Segment-EP*$(P^{t_i}, (t_{i+1} - t_i, \mathbf{e}_i^s, \mathcal{G}_{t_i, t_{i+1}})$
   3. Extract $P^{t_{i+1}}$ from the calibrated $\mathcal{G}_{t_i, t_{i+1}}$)
   4. Recalibrate $P^{t_{i+1}}$ and condition on $\mathbf{e}_{i+1}^p$

The last point addresses a subtlety relating to the propagation of messages from one interval to another. If a variable $X$ appears in two clusters $\mathcal{C}_i$ and $\mathcal{C}_j$ in a cluster graph, the distribution over its values in the two clusters is not generally the same, even if the EP computation converges. The reason is that even calibrated clusters only agree on the projected marginals over their sepset, not the true marginals. Thus, to obtain a coherent distribution $P^{t_{i+1}}$ to transmit to the next cluster graph, we should take the individual cluster marginals and sepsets for the state variables at time $t_{i+1}$, as obtained from $\mathcal{G}_{t_i,t_{i+1}}$, and recalibrate them to form a coherent distribution; the conditioning on point evidence can be done at the same time. We then extract $P^{t_{i+1}}$ as a set of calibrated cluster and sepset factors, and introduce each factor into the appropriate cluster or sepsent in $\mathcal{G}_{t_{i+2},t_{i+2}}$.

The algorithm *CTBN-Filter-EP* performs filtering — forward message passing. To perform smoothing, we can also pass messages in reverse, where the cluster graph for $[t_i, t_{i+1})$ passes a message to the cluster graph for $[t_{i-1}, t_i)$, representing the probability of the evidence after time $t_i$ given the state at $t_i$. Note that, to achieve more accurate beliefs, we can also repeat the forward-backward propagation until the entire network is calibrated, essentially treating the entire network as a single cluster graph. We omit details for lack of space.

Finally, we note that we chose to use one cluster graph for each segment of fixed continuous evidence. As a consequence, each cluster will approximate the trajectory of the variables it contains as a homogeneous Markov process, for the duration of the segment. We can modify the quality of the approximation by either refining or coarsening our choice of segments. In particular, if a set of variables is changing rapidly, we might want to partition a segment into subsegments, even if the evidence remains constant. Alternatively, we can reduce computational cost by collapsing several intervals of continuous evidence, approximating the trajectory distribution over the entire interval as a homogeneous Markov process. This step requires a more complex computation of sufficient statistics over the combined interval, but is not substantially different. The decision of how to partition time into intervals is analogous to a situation where we are approximating a distribution over continuous variables as a set of Gaussians, each defined over a subset of the space. The choice of how to partition the space into subsets determines the quality of our approximation.

### 5.3 Energy Functional

As for any EP algorithm over the exponential family, we can show that the convergence points of the EP algorithm in Sec. 5.1 are fixed points of the constrained optimization of the Kikuchi free energy functional, subject to calibration constraints on the projected marginals.

The Kikuchi free energy function for a cluster graph $\mathcal{G}$ is

$$\hat{\mathbf{F}}[P_\mathcal{N}, \hat{P}] = \quad (4)$$
$$\sum_{\phi \in \mathcal{N}} \mathbf{E}_{\pi_\phi}[\ln \phi] + \sum_{\mathcal{C}_i \in \mathcal{G}} \mathbf{H}_{\pi_i}(\mathcal{C}_i) - \sum_{\mathcal{C}_i - \mathcal{C}_j \in \mathcal{G}} \mathbf{H}_{\mu_{i,j}}(\mathcal{S}_{i,j})$$

subject to the constraints:

$$\mu_{i,j} = \mathrm{marg}_{\mathcal{C}_i \setminus \mathcal{S}_{i,j}}^{P^0, T}(\pi_i) \ . \quad (5)$$

**Theorem 5.2** *A set of potentials $\pi_i$, $\mu_{i,j}$ is a stationary point of maximizing Eq. (4) subject to Eq. (5) if and only if, for every edge $\mathcal{C}_i - \mathcal{C}_j$ there are potentials of the form*

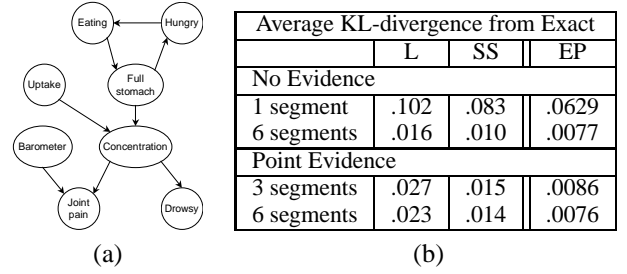

| Average KL-divergence from Exact | | | |
|---|---|---|---|
| | L | SS | EP |
| No Evidence | | | |
| 1 segment | .102 | .083 | .0629 |
| 6 segments | .016 | .010 | .0077 |
| Point Evidence | | | |
| 3 segments | .027 | .015 | .0086 |
| 6 segments | .023 | .014 | .0076 |

(a)                             (b)

Figure 1: (a)Drug effect network (b) Average KL-div. between the exact joint distribution and approximate distributions averaged over 60 time points.

$\delta_{i \to j}(\mathcal{S}_{i,j})$ *such that*

$$\delta_{i \to j} \propto \mathrm{marg}_{\mathcal{C}_i \setminus \mathcal{S}_{i,j}}^{P^0, T} \left( \pi_i^0 \times \prod_{k \in N_i - \{j\}} \delta_{k \to i} \right)$$
$$\pi_i \propto \pi_i^0 \times \prod_{j \in N_i} \delta_{j \to i}$$
$$\mu_{i,j} = \delta_{j \to i} \times \delta_{i \to j}$$

**Corollary 5.3** *Convergence points of the procedure* CTBN-Segment-EP *are stationary points of maximizing Eq. (4) subject to Eq. (5).*

The proof of these results is a special case of the general result on convergence of EP, which applies to any class of distributions in the exponential family.

## 6 Experimental Results

In our experiments, we used the drug effect network of NSK shown in Figure 1(a) allowing us to compare to the previous inference algorithm. We compared the results of our implementation of expectation propagation with exact inference and the approximate inference algorithm from NSK when possible. We ran three scenarios. In each one, at $t = 0$, the person modelled by the system experiences joint pain due to falling barometric pressure and takes the drug to alleviate the pain, is not eating, has an empty stomach, is not hungry, and is not drowsy. The drug is uptaking and the current concentration is 0. All scenarios ended at $t = 6$ (after 6 hours). We compare to exact inference by computing the average KL-divergence as discussed below.

In the first scenario, there was no evidence after the given initial distribution. We ran the algorithms viewing the entire trajectory as a single segment. We tried using one approximation to describe the dynamics over the system and also broke it down into 6 evenly spaced segments. In the second second scenario, we observe at $t = 1$ that the person is not hungry and at $t = 3$, that he is drowsy. We ran the algorithms with 3 segments and again with 6 segments.

NSK provide two approximate marginalizations: the linearization (L) and subsystem (SS) approximations. Also note that the NSK algorithms are single-pass multiply-marginalize instead of the multiply-marginalize-divide scheme of the EP algorithm. Figure 1(b) shows the average KL-divergence between exact joint distribution and the approximate joint distributions averaged over 60 evenly spaced time points between $t = 0$ and $t = 6$ for the experiments described above. From the table, one can see the expectation propagation easily beats the previous algorithms.

In the third scenario, we have continuous observations over the variables representing hunger, eating, and drowsiness. After the initial distribution given above, these three variables persisted in their initial state until $t = 0.5$, after which the person became hungry. At $t = 1$ the person begins to eat. At $t = 1.5$ the hunger is gone and at $t = 2$ the person stops eating. At $t = 2.5$ the person becomes drowsy and these three variables maintain their final value to the end of the trajectory at $t = 6$. We ran the EP forward filter with one segment for each interval of continuous evidence — a total of 6 segments (not evenly spaced). We again measured the average KL-divergence between the actual and approximate joint distributions as above, measuring at 60 evenly spaced time points between $t = 0$ and $t = 6$. The average KL-divergence was 0.00122. Allowing EP to run for only a single pass instead of going until convergence had a negligible effect — worsening the average KL-divergence by $6.7 \times 10^{-7}$. This is not surprising, as we found EP to converge rapidly: of the 6 segments we ran for the continuous evidence, all but one converged within a single pass.

## 7 Discussion and Conclusions

We have presented a new, well-founded, approximate inference algorithm for CTBNs that, for the first time, allows us to answer a full range of queries including the ability to handle continuous observations. We provided a view of CIM parametrization that enables cluster graph message passing algorithms that include division. Furthermore, we showed how we can compute a KL-divergence minimizing approximate marginalization of the distribution parametrized by the CIM.

These results enabled us to provide an expectation propagation algorithm for CTBNs which, subject to our approximate marginalizations, converges to stationary points of the approximate free energy function. Other approaches to approximate inference, such as sampling based methods, are ongoing work.

One of the most appealing properties of the algorithm that we proposed in this paper is that it adaptively selects the time granularity used for reasoning about a cluster based on the rate at which the cluster evolves. Different clusters will be discretized at different granularities, automatically selected by the integration algorithm. The same cluster may be discretized at one granularity in one interval of continuous evidence, and differently in another. By contrast, in DBNs, all variables in the system must be modeled at the time granularity of the variable that evolves most quickly. We can hope to extend this property further, by allowing one cluster in our network to cover a long interval, whereas another (over a different subset of variables) is partitioned into smaller segments. This could provide the basis for an algorithm that automatically and flexibly assigns computational resources to the parts of the system where the most interesting changes are occurring.

**Acknowledgments.** This work was funded by DARPA's EPCA program, under subcontract to SRI International, and by the Boeing Corporation.